\documentclass[10pt,twocolumn,letterpaper]{article}

\usepackage[pagenumbers]{cvpr} 

\usepackage{tcolorbox}
\usepackage{listings}
\usepackage{multirow}
\usepackage{graphicx}
\usepackage{tikz}
\usetikzlibrary{calc,positioning,arrows.meta}

\definecolor{cvprblue}{rgb}{0.21,0.49,0.74}
\usepackage[pagebackref,breaklinks,colorlinks,allcolors=cvprblue]{hyperref}

\def\paperID{**} 
\def\confName{CVPR}
\def\confYear{2026}
\def\papername{Plan-X}
\title{\papername: Instruct Video Generation via Semantic Planning}
\author{
Lun Huang$^{1,2,3}$\thanks{Work done during an internship at ByteDance.},
You Xie$^3$,
Hongyi Xu$^3$,
Tianpei Gu$^3$,
Chenxu Zhang$^3$,
Guoxian Song$^3$,\\
Zenan Li$^3$,
Xiaochen Zhao$^3$,
Linjie Luo$^3$,
Guillermo Sapiro$^{2,4}$
\vspace{2mm}\\
$^1$Duke University \quad
$^2$Princeton University \quad
$^3$ByteDance Intelligent Creation \quad
$^4$Apple\\
\url{https://byteaigc.github.io/Plan-X}\\
{\tt\small lun.huang@duke.edu} \quad
{\tt\small guillermos@princeton.edu} \\
{\tt\scriptsize \{you.xie, hongyixu, tianpei.gu, chenxuzhang, guoxiansong, zenan.li, xiaochen.zhao, linjie.luo\}@bytedance.com}
}

\begin{document}
\maketitle
\begin{abstract}
Diffusion Transformers have demonstrated remarkable capabilities in visual synthesis, yet they often struggle with high-level semantic reasoning and long-horizon planning. This limitation frequently leads to visual hallucinations and mis-alignments with user instructions, especially in scenarios involving complex scene understanding, human–object interactions, multi-stage actions, and in-context motion reasoning. To address these challenges, we propose \papername, a framework that explicitly enforces high-level semantic planning to instruct video generation process. At its core lies a Semantic Planner, a learnable multimodal language model that reasons over the user’s intent from both text prompts and visual context, and autoregressively generates a sequence of text-grounded spatio-temporal semantic tokens. These semantic tokens, complementary to high-level text prompt guidance, serve as structured ``semantic sketches'' over time for the video diffusion model, which has its strength at synthesizing high-fidelity visual details.
\papername\ effectively integrates the strength of language models in multimodal in-context reasoning and planning, together with the strength of diffusion models in photorealistic video synthesis. Extensive experiments demonstrate that our framework substantially reduces visual hallucinations and enables fine-grained, instruction-aligned video generation consistent with multimodal context.
\end{abstract}    
\section{Introduction}
\label{sec:intro}

Recent advances in Diffusion Transformers (DiTs)~\cite{peebles2023scalablediffusionmodelstransformers} have revolutionized visual generation, achieving unprecedented fidelity and temporal consistency in both image and video synthesis~\cite{sora,veo3,wan2.1,qiu2025skyreels,kong2024hunyuanvideo,gao2025seedance}.
Despite these successes, DiTs remain fundamentally limited in high-level semantic reasoning.
When prompted with complex or compositional instructions, such as multi-stage actions or human–object interactions within intricate scenes, they often produce visually plausible yet semantically inconsistent results.
Such failures typically manifest as prompt misalignment, visual hallucination, incorrect action ordering, or incoherent spatio-temporal relations, stemming from the model’s inability to jointly reason over textual guidance and visual context or to plan semantic evolution across extended temporal horizons.

The key difficulty of the problem lies in the entanglement between semantic reasoning and pixel-level synthesis.
Conventional video diffusion models are optimized to capture fine-grained spatial details and temporal dynamics but lack an explicit mechanism for semantic abstraction, that is, understanding how intentions, actions, and spatial relationships correlate with multimodal instructional context and evolve over time.
To alleviate this limitation, recent state-of-the-art video generation approaches~\cite{wan2.1,kuaishoukling,gao2025seedance} often incorporate auxiliary language models for prompt enhancement (PE), expanding high-level user prompts into detailed textual descriptions that better align with the training distribution.
While such text conditioning provides coarse global control, it still fails to enforce structured, frame-level semantic consistency, often resulting in undesirable hallucinations or semantic drift.
Moreover, even with detailed and accurate textual guidance, the underlying DiT may still struggle to interpret complex multimodal instructions, particularly in long-horizon or compositional scenarios.

In this work, we argue that semantic reasoning and visual synthesis should be treated as distinct yet complementary processes.
We introduce \papername, a framework that decouples high-level instruction understanding and semantic planning from holistic video synthesis.
The central idea is to offload semantic reasoning to a trainable multimodal language model, termed the Semantic Planner, which interprets the user’s prompt and visual context and autoregressively generates a sequence of spatio-temporal semantic visual tokens.
Unlike abstract textual scripting, the Semantic Planner is trained to produce spatially-grounded visual tokens corresponding to sequential keyframes, serving as structured ``semantic sketches'' describing what should happen and when.
These structured semantic sketches then guide a video DiT, which focuses solely on decoding the semantic sketches into high-fidelity, temporally consistent videos.
To bridge textual and visual semantics, we adopt text-aligned tokens (TA-Tok)~\cite{han2025tar}, derived from quantized SigLIP2~\cite{tschannen2025siglip2multilingualvisionlanguage} representations via a language-model codebook.
This design enables the language model to translate high-level multimodal instructions into spatio-temporal visual structures over an extended multimodal vocabulary, providing interpretable and controllable conditioning for video synthesis.

This design offers two key advantages.
First, it transforms the challenge of joint semantic understanding and generation into a language modeling problem, enabling efficient in-context multimodal reasoning and compositional planning through text-aligned semantic token representations.
Second, it preserves the strengths of diffusion transformers in realistic, temporally coherent rendering without overburdening them with abstract reasoning.
Together, these components form a unified multimodal system capable of interpreting, planning, and executing complex video generation tasks.

We extensively evaluate \papername~on a challenging video generation benchmark covering text-to-video, image-to-video, and video continuation tasks.
Across all settings, \papername~achieves substantial improvements, both quantitatively and qualitatively, in prompt alignment, semantic coherence, and motion naturalness across diverse scenarios, including human–object interaction, multi-stage actions, and long-horizon planning.
Furthermore, our experiments show that the proposed Semantic Planner not only mitigates visual hallucination but also enables interpretable and transferable semantic guidance, facilitating downstream applications such as semantic cross-transfer.
\begin{figure*}[t!]
\centering
 \includegraphics[width=\linewidth]{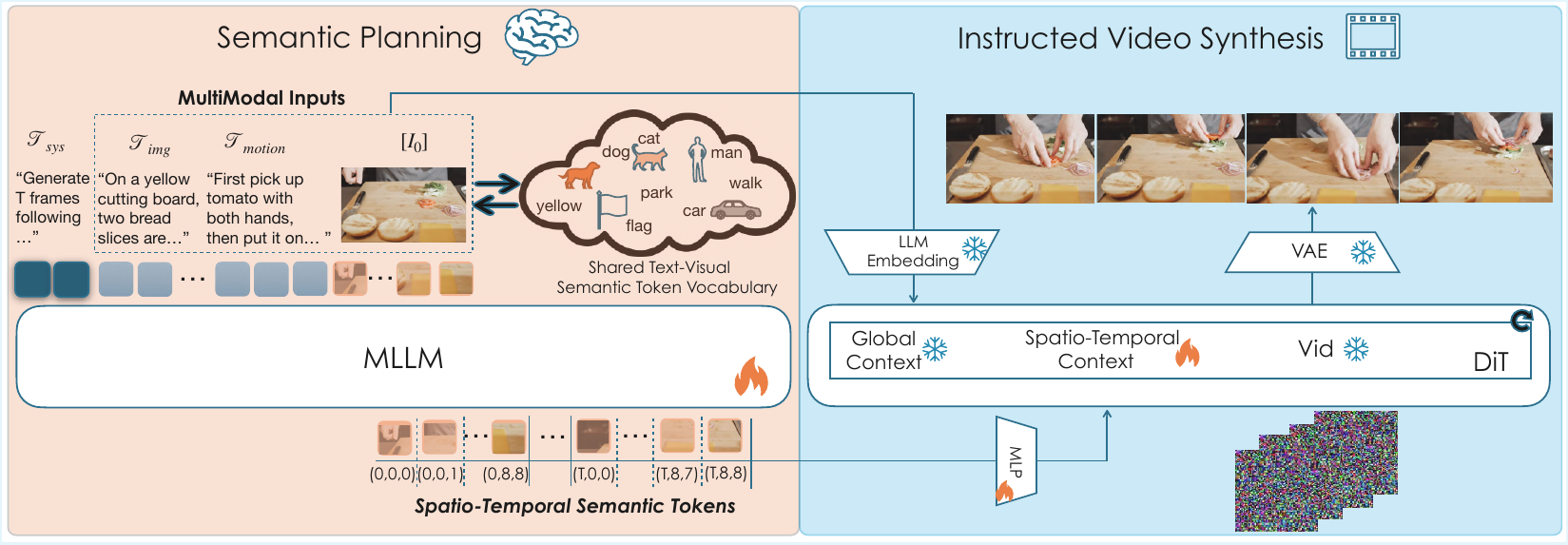}
   \caption{ 
   \textbf{Overview of \papername.} \papername~comprises two key components: an MLLM for high-level semantic reasoning and planning, and a DiT for high-fidelity video synthesis. The Semantic Planner receives multimodal inputs, including a descriptive text prompt for static scene content ($\mathcal{T}_{\text{img}}$), a motion description ($\mathcal{T}_{\text{motion}}$), an instructional system prompt ($\mathcal{T}_{\text{sys}}$) specifying the number of target frames $T$, and optionally the semantically encoded first frame $I_0$.
It then autoregressively generates discrete text-aligned semantic tokens that encode spatio-temporal semantic structures in the form of keyframes.
Complementary to global text conditioning, we introduce a dedicated semantic guidance branch that instructs a pretrained DiT model to translate these structured semantics augmented with 3D spatio-temporal RoPE into high-fidelity, temporally coherent video realizations.
   }
   \vspace{-10pt}
    \label{fig:pipeline}
\end{figure*}

\section{Related Works}

\paragraph{Video Generation.} 
Fueled by rapid advances in large-scale generative modeling, both closed- and open-source video generation systems~\cite{sora,kuaishoukling,wan2025wan,gao2025seedance,kong2024hunyuanvideo,veo3} have made remarkable progress, particularly within diffusion-based frameworks.
Early architectures primarily adopted U-Net backbones~\cite{DBLP:journals/corr/RonnebergerFB15} initialized from image diffusion models~\cite{ho2022videodiffusionmodels}, later extended to the video domain through temporal modeling mechanisms~\cite{guo2023animatediff}.
Recent research has shifted toward Diffusion Transformers (DiTs)~\cite{peebles2023scalablediffusionmodelstransformers}, which achieve substantially higher visual fidelity and temporal consistency than U-Net variants.
The original DiT architecture employs cross-attention to connect video and text embeddings, whereas MM-DiT frameworks~\cite{esser2024scaling,genmo2024mochi,shin2025exploringmultimodaldiffusiontransformers} concatenate text embeddings with visual tokens for full-sequence attention.
The text encoder plays a central role in determining semantic alignment: current state-of-the-art systems often utilize large language models (LLMs) such as T5~\cite{raffel2023exploringlimitstransferlearning} or Qwen~\cite{team2024qwen2}, sometimes coupled with CLIP~\cite{radford2021learningtransferablevisualmodels} to strengthen multimodal consistency.
Moreover, to align user prompts with caption-style conditioning distributions, many frameworks incorporate an auxiliary language model, often termed a Prompt Enhancer~\cite{Yang2024Qwen25TR}, that expands or reformulates user instructions before synthesis.
Despite these advancements, existing video diffusion models still exhibit semantic misalignment and visual hallucination when handling complex, multi-action, or long-horizon prompts, due to the lack of explicit semantic reasoning and temporal planning mechanisms.

\vspace{-3mm}
\paragraph{Unified MultiModal Visual Generation.}
Beyond text-conditioned generation, recent research has increasingly focused on unifying multimodal understanding and generation within a single framework~\cite{xie2024showo,wu2024vila,qu2025tokenflow,yao2025reconstruction,kim2025democratizing,han2025tar,pan2025transfer,ma2025janusflow,sun2024generativemultimodalmodelsincontext,chen2025blip3ofamilyfullyopen,zhou2024transfusion,lin2025bifrost,deng2025emerging,tong2024metamorph,liao2025mogao,wang2025selftok,wu2025qwen,xie2025xstreamerunifiedhumanworld,zheng2025diffusion}.
To enable cross-modal generation, many of these approaches develop semantics-aware, text-aligned visual encoders~\cite{wang2025selftok,wang2024emu3,team2024chameleon,kim2025democratizing,yao2025reconstruction}, adapt pretrained semantic encoders for pixel representation~\cite{wu2024vila,lin2025bifrost,wu2025qwen,han2025tar}, or combine both strategies in hybrid fashion~\cite{deng2025emerging,ma2025janusflow}.
TA-Tok~\cite{han2025tar} exemplifies this direction by converting SigLIP2 embeddings~\cite{tschannen2025siglip2multilingualvisionlanguage} into discrete text-aligned tokens, forming a unified token space that bridges textual and visual modalities.
In our work, we leverage this tokenizer to represent both video frames and text prompts within a shared semantic space, facilitating fine-grained cross-modal reasoning.
To further harness the multimodal reasoning capacity of multimodal large language models (MLLMs), recent works either train them for autoregressive prediction of visual tokens~\cite{chen2025blip3onextfrontiernativeimage,xie2024showo,zhou2024transfusion,deng2025emerging} or introduce learnable query encodings~\cite{ge2024seed,pan2025transfer,chen2025blip3ofamilyfullyopen}, sometimes paired with diffusion heads for high-fidelity image synthesis in continuous latent spaces.
However, these approaches primarily target static image generation or editing, and thus remain limited in modeling temporal dynamics and long-range semantic reasoning.
\papername~extends this generative paradigm to the video domain by bridging MLLMs and video DiTs for high-fidelity, temporally consistent video synthesis under multimodal semantic reasoning and planning.

\section{Method}


Recent video diffusion models~\cite{wan2025wan,gao2025seedance,kong2024hunyuanvideo,seawead2025seaweed}, denoted as $\mathcal{G}$, are typically built upon Diffusion Transformers (DiTs)~\cite{peebles2023scalableDiT} and generate videos $\mathcal{V}$ conditioned on a text prompt $\mathcal{T}$, optionally combined with a reference image $I_0$ serving as the first frame.
These models are generally trained with denoising objectives, emphasizing pixel-level fidelity and temporal smoothness.
However, despite their impressive visual quality, diffusion models often struggle with high-level semantic interpretation and reasoning, as well as maintaining coherent semantic structure and evolution over extended temporal horizons.
In contrast, large language models (LLMs)~\cite{comanici2025gemini,touvron2023llama,guo2025deepseek,Yang2024Qwen25TR,openai2025gpt5} excel at abstract, high-level semantic understanding and long-horizon reasoning, but they still lag behind diffusion models in video synthesis due to error accumulation, coarse spatial grounding, and limited visual expressiveness.

To overcome these limitations, we propose a decoupled design that separates high-level semantic reasoning and planning from low-level visual synthesis.
Specifically, instead of merging semantic planning and video generation into a single DiT model via
$\mathcal{V} = \mathcal{G}(\mathcal{T}, I_0)$,
our approach delegates complex semantic reasoning to a specialized multimodal large language model $\mathcal{M}$, which transforms multimodal prompts into spatio-temporal semantic guidance.
The diffusion transformer $\mathcal{G}$ is then conditioned on both the original inputs and the generated semantic blueprint $\mathcal{S}$ to perform semantics-driven video synthesis:
\vspace{-3mm}
\begin{equation}
\mathcal{S} = \mathcal{M}(\mathcal{T}, I_0),
\quad\quad
\mathcal{V} = \mathcal{G}(\mathcal{S}, \mathcal{T}, I_0).
\vspace{-3mm}
\end{equation}

This synergistic modular framework enables each component to specialize:
the multimodal language model $\mathcal{M}$ captures and structures semantic intent and temporal logic, while the diffusion transformer $\mathcal{G}$ focuses on realizing these semantics through high-fidelity, temporally consistent synthesis. 
An overview of the proposed framework is illustrated in Figure~\ref{fig:pipeline}.

\subsection{Text-Grounded Semantic Planning}
\label{sec:plan}
To bridge the language model and the video DiT through semantic scripting, we evaluate three potential design choices.
The most common approach is text based conditioning, where the language model serves as a prompt enhancer that expands high level and abstract user instructions into detailed textual descriptions.
While this representation is native to the language model’s capability, textual semantics remain difficult for the DiT to fully interpret and realize.
Moreover, when translating between concise prompts and detailed captions, the language model is prone to hallucination~\cite{llmhall2025}, which leads to semantic drift and prompt misalignment.
Another choice adopted in several unified frameworks for image understanding and generation~\cite{ge2024seed,pan2025transfer,chen2025blip3ofamilyfullyopen} employs implicit representations, such as hidden states from the language model or with fixed-length learnable queries, as conditioning inputs for downstream diffusion models.
However, these implicit embeddings lack explicit spatio-temporal structure and are not flexible representing videos with varying resolution or duration.
As a result, the diffusion model remains overburdened with translating abstract global features into semantically coherent videos instead of focusing on high-fidelity visual synthesis.

Therefore, we adopt a bridging design that emphasizes high-level spatio-temporal visual semantics across variable duration, while remaining interpretable by both the language model and the diffusion model.
To achieve this, our semantic planner functions as an autoregressive model over text-grounded visual semantic tokens, generating semantic trajectory that directs the video diffusion.





\vspace{-3mm}
\paragraph{Spatio-Temporal Semantic Tokens.}
Visual representation encoders such as CLIP~\cite{radford2021learningtransferablevisualmodels}, SigLIP2~\cite{tschannen2025siglip2multilingualvisionlanguage}, and DINO~\cite{oquab2023dinov2} have been widely adopted in visual understanding tasks, demonstrating strong structural and semantic representation capability.
Recent studies~\cite{zheng2025diffusion,shi2025latentdiffusionmodelvariational} have further shown that their latent spaces can support both visual reconstruction and generation.
Motivated by these findings, we leverage the semantic latent space of the \emph{SigLIP2-so400m-patch14-384}, where the spatial structure and temporal evolution of a video are represented through uniformly sampled keyframes that are patchified into tokens.
SigLIP2 tokens are trained to capture both global and local semantic structures while remaining aligned with textual representations.
These text-grounded semantic tokens are interpretable by the language model and simultaneously provide semantically rich spatio-temporal structure for the video DiT.

In contrast to abstract textual descriptions that serve only as global conditions, our semantic tokens provide concrete structural sketches across both spatial and temporal dimensions.
To avoid overburdening the language model $\mathcal{M}$ with pixel-level details or dense frame-wise temporal modeling, we sample sparse keyframes at 2 FPS.
Each keyframe is uniformly resized to $384 \times 384$ and encoded into patchified spatial tokens.
To further facilitate semantic token generation within the language model, we employ TA-Tok~\cite{han2025tar}, which discretizes SigLIP2 embeddings using a pretrained LLM-based text codebook.
Specifically, TA-Tok fine-tunes the SigLIP2 encoder and quantizes its visual embeddings through a text-aligned codebook initialized from LLM embeddings, forming an expanded visual vocabulary.
Notably, we apply scale-adaptive pooling to obtain $N=81$ spatial tokens per keyframe, denoted as $\mathcal{S}$, achieving a balanced abstraction between high-level semantics and spatial precision.
This visual–text token representation enables the language model to perform cross-modal reasoning and generation within a shared discrete semantic space, effectively bridging visual and linguistic modalities.


\paragraph{Autoregressive Semantic Planning}

To accommodate semantic planning over adaptive temporal durations, we formulate the task as a next-token prediction problem, where a LLM autoregressively generates spatio-temporal semantic tokens $\mathcal{S}$ given multimodal context inputs.
Specifically, we adopt Qwen-2.5-Instruct~\cite{Yang2024Qwen25TR} as the LLM backbone $\mathcal{M}$, which autoregressively predicts a sequence of semantic keyframes $\mathcal{S}_t$ at $2$ FPS, conditioned on the textual prompt $\mathcal{T}$ and optionally a semantically encoded initial frame $\mathcal{S}_0$.

We standardize all textual prompts $\mathcal{T}$ into a unified instruction-following format, which describes both the static scene content $\mathcal{T}_{img}$ and the intended video dynamics $\mathcal{T}_{motion}$, together with the target duration specified as $T$ keyframes $\mathcal{T}_{sys}$.
Our training strategy supports multi-task generation within a single unified framework, encompassing text-to-video, image-to-video, and video continuation tasks.
Notably, the language model $\mathcal{M}$ jointly interprets both textual and visual context (for example, the first frame) within a shared semantic token space, thereby strengthening the correspondence between visual content and user instructions.
This design effectively mitigates visual hallucination that often arises from conventional prompt enhancement methods.
We employ fully causal attention across all semantic tokens, enabling streamable generation with flexible and extensible temporal durations.
The language model $\mathcal{M}$ is trained to autoregressively predict the next semantic tokens by minimizing the following objective:

\vspace{-4mm}
\begin{equation}
    \mathcal{L}_{\text{planner}} = \sum_{t=1}^{T} \sum_{i=1}^{N} -\log P(\mathcal{S}_{t,i} | \mathcal{S}_0, \mathcal{T}, \mathcal{S}_{<t}, \mathcal{S}_{t,<i})
    \label{eq:planner_loss}
\end{equation}
\vspace{-2mm}
where $\mathcal{S}_{t,i}$ is the $i$-th spatial token of the $t$-th frame.




\subsection{Semantics-Instructed Video Synthesis}
Diffusion models have been widely adopted for video synthesis, where the video latent $\mathcal{V}$ is learned to progressively denoise from a noise tensor $\mathbf{z} \sim \mathcal{N}(0, \mathbf{I})$.
The text caption serves as a global contextual condition throughout the denoising process, typically incorporated through cross-attention mechanisms or within a multimodal DiT (MMDiT) architecture.
In this section, we describe how the synthesized spatio-temporal semantics $\mathcal{S}$ are integrated into a pre-trained DiT-based video diffusion model $\mathcal{G}$, while preserving the model’s intrinsic video generation capabilities in visual fidelity and temporal coherence.



\vspace{-3mm}
\paragraph{In-Context Semantics Guidance.}
To achieve this integration, we replace the original text-conditioning branch in $\mathcal{G}$ with a hybrid conditioning module that attends video latents to both the high-level text prompt $\mathcal{T}$ for global context and the spatio-temporal semantics $\mathcal{S}$ for fine-grained structural guidance.
Specifically, we replicate the text conditioning pathway for $\mathcal{S}$, either through duplicated cross-attention layers or by adding a dedicated semantic modality branch within the MMDiT architecture, depending on the design of the underlying diffusion model. The semantic token embeddings are projected through a lightweight two-layer MLP, producing semantic context features that are aligned with the DiT hidden-state dimensionality. 

In contrast to channel-wise structural offsets commonly used in ControlNet~\cite{zhang2023adding} or PoseGuider~\cite{hu2024animate}, our approach treats $\mathcal{S}$ as a complementary semantic context that spans both spatial and temporal dimensions, rather than a deterministic control signal.
This design allows $\mathcal{G}$ to fully exploit its inherent capability for pixel-level visual realization, while remaining robust against misalignment or inconsistency propagated from the semantic planner. 

However, unlike textual context conditioning that spans the entire spatio-temporal spectrum, each semantic token encodes a localized context, specifying what occurs, where, and when.
To strengthen the alignment between semantic tokens and video latents, we enrich $\mathcal{S}$ with 3D time-aligned spatio-temporal Rotary Position Embeddings (RoPE), introducing explicit positional correspondence across both space and time.
Let $\mathbf{q}_v$ denote the query tensor derived from the video latents $\mathcal{V}$ and $\mathbf{k}_s$ the key tensor derived from the semantic tokens $\mathcal{S}$.
The rotary-encoded query $\tilde{\mathbf{q}}_v$ and key $\tilde{\mathbf{k}}_s$ are computed as

\vspace{-6mm}
\begin{equation}
\tilde{\mathbf{q}}_{v,i} = R(\theta, p_{v,i}) \mathbf{q}_{v,i},
\quad\quad
\tilde{\mathbf{k}}_{s,j} = R(\theta, p_{s,j}), \mathbf{k}_{s,j},
\end{equation}
where $R(\theta, p)$ is the rotation matrix parameterized by base frequency $\theta$ at position $p$.
The position indices $p_{v,i} = (t, h, w)$ and $p_{s,j} = (t', h', w')$ correspond to the spatio-temporal coordinates in the video latent and semantic token grids, respectively.
Notably, the semantic tokens and video latents have different compression ratios across spatial resolutions, and the semantic tokens are sampled more sparsely in time.
Accordingly, RoPE frequencies are scaled based on the 3D grid resolution, ensuring that each position in the semantic map maintains consistent relative alignment with its corresponding location in the video latent feature space.

\vspace{-3mm}
\paragraph{Staged Training.}
The training of the video generator $\mathcal{G}$ is conducted in two stages.
We initialize the weights of the newly added semantic context branch using those of the text branch, and in the first stage, only the new semantic branch is trained while the original model parameters are frozen to preserve the pretrained generative capability.
Specifically, we encode the ground-truth video into semantic tokens using the TA-Tok tokenizer, and train the model with the native flow-matching loss that reconstructs the velocity field $\mathbf{v}$ for denoising video latents $\mathbf{x}_0$ from noise $\mathbf{z}$:

\vspace{-4mm}
\begin{equation}
\mathcal{L}_{\text{vid}}
=\mathbb{E}_{\mathbf{x}_0, \mathbf{z}, k \sim \pi(k)}\left[
\| \mathbf{v} - \mathbf{v}_\theta(\mathbf{x}_k, k, \mathcal{C}) \|_2^2
\right],
\end{equation}
where $\pi(k)$ is a sampling distribution for timestep $k$, and $\mathcal{C}$ denotes the conditioning set that may include both text and semantic guidance.
At the beginning of training, we fully drop the text conditioning, enforcing the diffusion model to rely solely on the semantic context.
As training progresses, the text drop rate is gradually reduced to 0.5, allowing the model to restore its text-conditioned generation capability while learning to co-utilize both text and semantic guidance.

To further improve robustness to imperfect semantic predictions, we fine-tune the model end-to-end by replacing ground-truth semantic tokens with hidden states of generated tokens from the semantic planner.
This stage mitigates exposure bias and enhances the model’s self-adaptation to noisy semantic inputs during inference.

\begin{figure*}[t]
	\centering
	\includegraphics[width=0.99\textwidth]{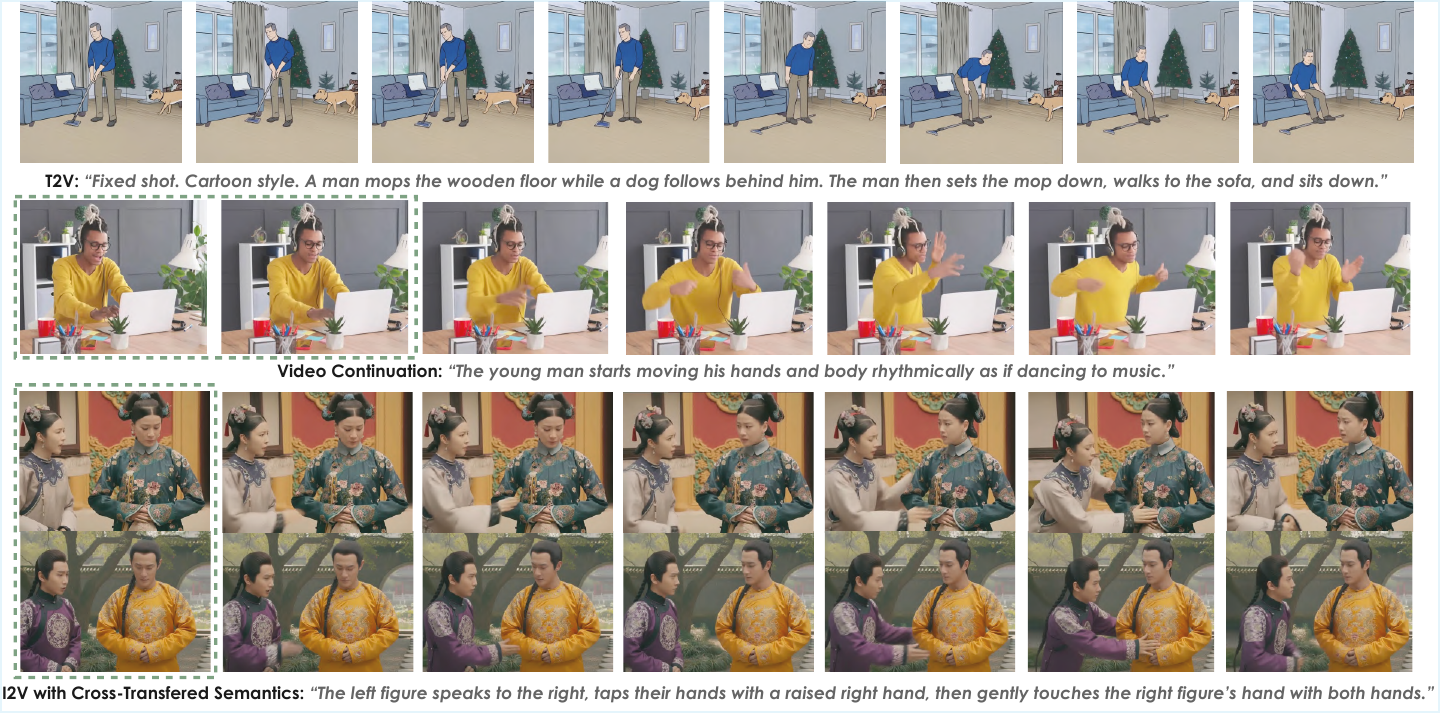}
    \vspace{-2mm}
	\caption{\textbf{Qualitative results.} \papername~produces high-fidelity, semantically consistent, and instruction-aligned videos across T2V, I2V, video continuation, and semantic transfer, all within a single unified framework.}
    \vspace{-5mm}
	\label{fig:applications}
\end{figure*}

\section{Experiments}
\subsection{Implementation Details}

\paragraph{Dataset.} We curate a large-scale corpus of 4.5M training video clips by combining multiple public datasets and in-house collections.
Specifically, we include human–object interaction datasets such as Taste-Rob~\cite{Zhao_2025_CVPR} and HOIGen-1M~\cite{HOIGen}, along with a diverse general-domain video dataset collected internally.
For unified video captioning, we employ Qwen3-VL-32B~\cite{Qwen2.5-VL}, generating descriptions that capture both static scene content and dynamic motion. 

For evaluation, we construct a challenging benchmark comprising samples from the test split of Taste-Rob~\cite{Zhao_2025_CVPR} and additional in-the-wild cases, covering both text-to-video and image-to-video tasks.
The benchmark includes 50 test cases per task, featuring human–object interactions in complex scenes and multi-step action sequences.
In contrast to the prompt suite in VBench~\cite{huang2023vbench}, our benchmark specifically evaluates semantic planning capability under multi-context and compositional actions scenarios. 

\vspace{-6mm}
\paragraph{Training and Inference.} 
Our full pipeline includes variants of both the semantic planner and the video diffusion model.
The semantic planner is initialized from Qwen2.5-Instruct~\cite{Yang2024Qwen25TR} with either 1.5B or 7B parameters, while the video diffusion backbones are based on Wan 2.2 (5B)~\cite{wan2025wan} or Seedance 1.0 ~\cite{gao2025seedance}.
By default, we report results using the 7B semantic planner paired with Wan 2.2 (5B), and analyze the effects of model scale and backbone choice in the ablation studies presented later.
Training of the semantic planner and diffusion model is performed on 48 A100 GPUs with an effective batch size of 48, using the AdamW optimizer with learning rates of $5\times10^{-5}$ and $2\times10^{-5}$, respectively.
The semantic planner and diffusion model are trained for 7 epochs and 2 epochs, followed by joint end-to-end fine-tuning for one additional epoch.

Our semantic planner is trained with a context window of 4K tokens, supporting extensible generations of 20 seconds, and we use a sampling temperature of 0.9 during inference. The diffusion model is inferred with 50 denoising steps under a classifier-free guidance weight of 5.0. For unconditional inputs, we condition on the initial frame for the I2V task and use negative text prompts to suppress low-quality aesthetics and unnatural motion. Please refer to the supplementary materials for additional dynamic visual results and extended evaluations.

\subsection{Evaluation}
\paragraph{Baselines.}
We compare \papername~with five state-of-the-art video diffusion models, including Wan 2.2-5B~\cite{wan2025wan}, HunyuanVideo~\cite{kong2024hunyuanvideo}, and SkyReelsV2-14B~\cite{qiu2025skyreels}, as well as two commercial systems, Kling 1.6~\cite{kuaishoukling} and Seedance 1.0~\cite{gao2025seedance}.
Comprehensive comparative experiments are conducted across these baselines and our approach under the proposed I2V benchmark.
All baseline models are evaluated using their default inference settings, without additional fine-tuning and super-resolution.
For consistency, all generated videos are resized to 480p for fair comparison.

\begin{table*}[ht]
	\centering
	\small
	\setlength{\tabcolsep}{7pt}
	\renewcommand{\arraystretch}{1.15}
    \vspace{-2mm}
	\caption{\textbf{Quantitative evaluation and ablation.} We denote the best results in \textbf{bold} and the second-best results with \underline{underline}. }
	\label{tab:quan-main}
    \resizebox{\textwidth}{!}{
	\begin{tabular}{lccccccc}
		\toprule
Method  & Accuracy $\uparrow$ & Completeness$\uparrow$ & Fidelity $\uparrow$ & Consistency $\uparrow$ & Naturalness $\uparrow$ & Visual $\uparrow$ & Human Pref.$\uparrow$ \\
\midrule
Wan 2.2-5B~\cite{wan2025wan}   & 0.3156 & 0.3967 & 0.4389 & 0.5178 & 0.8378 & 0.8778 & 0.164  \\

HunyuanVideo~\cite{kong2024hunyuanvideo}  & 0.4467 & 0.5133 & 0.5933 & 0.9089 & 0.8889 & 0.9178 & 0.070 \\

SkyReelsV2-14B~\cite{qiu2025skyreels}  & 0.5575 & 0.5775 & 0.7725 & 0.9100 & 0.8500 & 0.9400 & 0.110 \\

Kling 1.6~\cite{kuaishoukling}   & 0.6810 & 0.7357 & 0.8119 & 0.9095 & 0.8976 & 0.9381 &  0.176 \\

Seedance 1.0~\cite{gao2025seedance}  & {0.7114} & {0.7943} & 0.8818 & 0.9455 & {0.9205} & {0.9500} & \underline{0.218} \\
\midrule

Wan 2.2 5B + SFT   & 0.5628 & 0.6651 & 0.7488 & 0.8814 & 0.8605 & 0.9186 &  -\\

Wan 2.2 5B + query   & 0.7030 & 0.7576 & 0.8576 & 0.9455 & 0.8939 & 0.9303 & - \\

Wan 2.2 5B + PE    & 0.5773 & 0.6795 & 0.7977 & 0.9136 & 0.8568 & 0.9364 & - \\

\papername{}-Wan w/o 3D RoPE   & 0.5889 & 0.6689 & 0.6911 & 0.7311 & 0.6844 & 0.7333 & - \\


\papername{}-Wan (1.5B $\mathcal{M}$)    & {0.7556} & {0.8111} & {0.9481} & \textbf{0.9852} & {0.9111} & {0.9519} & - \\
\midrule
\papername{}-Wan   & \underline{0.7816} & \underline{0.8263} & \underline{0.9500} & \underline{0.9816} & \textbf{0.9394} & \textbf{0.9737} & \textbf{0.262} \\

\papername{}-Seedance  & \textbf{0.7971} & \textbf{0.8571} & \textbf{0.9657} & {0.9657} & \underline{0.9343} & \underline{0.9629} &  -\\
\bottomrule
\end{tabular}
}
\vspace{-5mm}
\end{table*}

\begin{figure*}[t]
	\centering
	\includegraphics[width=0.99\textwidth]{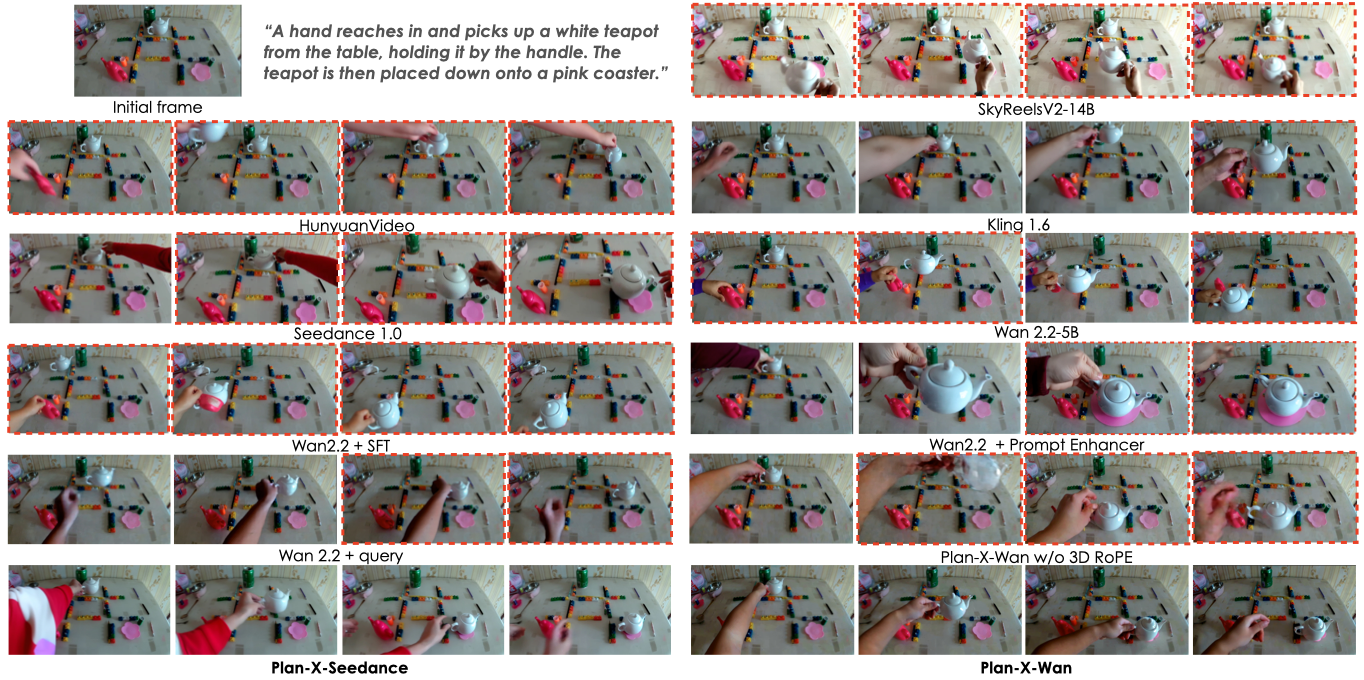}
	\caption{\textbf{Qualitative comparison and ablation.} Enabled by our semantic planner, both \papername-Wan and \papername-Seedance jointly reason over the provided visual content and the user’s text instruction, producing actions that accurately follow the intended semantics. In contrast, baseline methods frequently deviate from the instruction, exhibiting visual hallucination (SkyReelsV2, HunyuanVideo), interacting with misaligned objects or incorrect spatial locations (Seedance, Wan2.2), or producing severe object misplacement (Wan2.2, Wan2.2-SFT, Kling). Ablated variants that replace our semantic planner with text-based prompt enhancement or query-based conditioning also suffer from prompt misalignment.}
	\label{fig:results_comparison}
    \vspace{-4mm}
\end{figure*}

\vspace{-4mm}
\paragraph{Qualitative Evaluation.}
In Figure~\ref{fig:teaser} and~\ref{fig:applications}, we showcase the capability of \papername~in synthesizing high-fidelity, prompt-aligned videos through multimodal reasoning and multi-stage semantic planning.
Specifically, we illustrate:
the \textbf{(1) multimodal planning } ability of the semantic planner, which jointly reasons over all input modalities in a unified manner and generates coherent semantic trajectories across combinations of contexts, such as text, first frame, and initial video chunk, manifesting as
T2V, TI2V and video continuation tasks. 
Notably, we highlight the model’s strength in handling \textbf{(2) human object interactions}, where accurate alignment between visual content and textual instruction is essential to eliminating visual hallucination.
For example, in the task “pick up an object and move it to a target location,” \papername~leverages spatial reasoning and grounding from the provided image to ensure semantic and visual consistency, whereas prior approaches often hallucinate new or misplaced objects due to insufficient coupling between textual intent and visual context.
We further demonstrate the superiority of our semantic planner in \textbf{(3) long-range, multi-action planning}, where complex, temporally ordered actions are executed smoothly and coherently, unlike prior approaches that frequently omit actions or generate them in incorrect order. 
Lastly, we demonstrate a side application of our framework in \textbf{(4) semantics transfer}, where the generated semantic tokens can be seamlessly applied to different initial frames, since they encode high-level semantics and are injected into the DiT as contextual guidance rather than spatially aligned offsets. Visual comparisons between \papername~and baseline methods are presented in Figure~\ref{fig:results_comparison}, highlighting clear advantages in prompt alignment and semantic consistency.

\vspace{-6mm}
\paragraph{Quantitative Evaluation.}
There are currently no reliable numerical metrics for assessing interaction accuracy, prompt alignment, or motion fidelity in complex compositional video generation. To address this, we leverage the state-of-the-art MLLM Gemini 2.5~\cite{comanici2025gemini} to score model outputs on a normalized scale from 0 (worst) to 1 (best). Specifically, we evaluate human–object interaction accuracy (Accuracy$\uparrow$) and fidelity (Fidelity$\uparrow$), temporal motion completeness (Completeness$\uparrow$), scene consistency (Consistency$\uparrow$), motion naturalness (Naturalness$\uparrow$), and overall visual quality (Visual$\uparrow$). The full set of evaluation prompts used for MLLM scoring is provided in the supplementary materials. In addition, we conduct a human preference study with 16 participants, who are asked to select the two best videos between \papername-Wan and each competing baseline based on overall instruction alignment.

As shown in Table~\ref{tab:quan-main}, \papername{} achieves the highest performance on nearly all MLLM-evaluated dimensions. Compared to strong baselines such as Seedance 1.0 and Kling 1.6, our method yields substantial gains in compositional alignment, with Accuracy increasing from 0.7114 (Seedance) to 0.7971 and Completeness from 0.7943 to 0.8571. Moreover, \papername{} exhibits minimal visual hallucination (Fidelity$\uparrow$, Consistency$\uparrow$), outperforming or matching the strongest existing models in both dimensions. In terms of visual stability, \papername-Wan matches or surpasses all baselines, reaching the top Visual score of 0.9737 and the highest Naturalness score of 0.9394.

Further, \papername-Wan shows strong improvements over its pretrained backbone (Wan 2.2-5B), outperforming it by +0.4660 Accuracy, +0.4296 Completeness, and +0.5111 Fidelity, demonstrating the effectiveness of our proposed components. The superiority of \papername{} is corroborated by human preference results, where \papername-Wan achieves the highest preference score of 0.262, outperforming all competing methods, including the previously strongest baseline Seedance 1.0 (0.218). These results confirm that \papername{} produces videos that are not only more aligned to the instructions but also perceptually more coherent and natural to human viewers.

\vspace{-6mm}
\paragraph{Ablations.}
To assess the impact of our semantic planner, we perform ablation studies using the same diffusion backbone, Wan 2.2-5B, and compare against three alternative variants:
\textbf{(1) SFT}, where the DiT model is directly fine-tuned on our training corpus for the same number of epochs, without any semantic planning;
\textbf{(2) Prompt Enhancement}, where a built-in prompt enhancer is used to rewrite user instructions before generation; and
\textbf{(3) Query-style} conditioning~\cite{pan2025transfer}, in which a Qwen2.5-VL-3B~\cite{Qwen2.5-VL} is frozen as an MLLM backbone and a learnable 256-token query, along with a 24-layer transformer connector, is trained to provide additional conditioning to the DiT.
As evidenced by the quantitative results in Table~\ref{tab:quan-main} and the qualitative comparisons in Figure~\ref{fig:results_comparison} as well as the supplementary material, our proposed semantic planner substantially outperforms all ablated variants, yielding significantly stronger prompt alignment while avoiding visual hallucinations.
These results demonstrate that explicit semantic planning offers more structured and interpretable guidance than both text-only and query-based conditionings.

We further ablate the effectiveness of 3D RoPE, which establishes spatio-temporal correspondence between semantic tokens and video latents, thereby improving prompt alignment.
Lastly, we ablate the impact of both the semantic planner model scale and the choice of pretrained DiT backbone.
Using the same training corpus, we train semantic planners with 1.5B and 7B parameters, where stronger semantic planning capability emerges at the larger model scale.
Our framework represents a general generative paradigm, and the semantic planner is compatible with diverse video diffusion architectures Wan2.2, which employs cross-attention for text conditioning, and Seedance 1.0, which adopts an MMDiT architecture.
As shown in Table~\ref{tab:quan-main}, our semantic planner consistently improves the prompt-following capability of each base DiT model without compromising pretrained visual fidelity. Additional ablation studies are provided in the supplementary material.

\section{Conclusion}

We introduced~\papername, a framework that decouples high-level semantic planning from low-level video synthesis.
A Semantic Planner, implemented as a multimodal language model, interprets user instructions and visual context into spatio-temporal semantic tokens, which guide a DiT model for high-fidelity video generation.
This separation of reasoning and rendering effectively bridges instruction understanding and visual synthesis, reducing hallucination and improving prompt alignment, semantic coherence, and controllability across diverse video generation tasks.
\papername~highlights a new video generation paradigm where language models act as interpretable planners, enabling diffusion transformers to generate videos that are both semantically grounded and visually precise.

\vspace{-4mm}
\paragraph{Future work.} While this work focuses primarily on video generation, our framework is inherently extensible to video understanding and editing within a unified architecture, which we plan to explore in future research.
Currently, we leverage a pretrained discrete semantic tokenizer from TA-Tok~\cite{han2025tar}, which exhibits limited expressiveness in representing complex concepts and abstract reasoning (such as mathematical or symbolic content).
In future work, we aim to develop more expressive text-aligned visual semantic tokenizers, enabling richer multimodal understanding and video synthesis.

\vspace{-3mm}
\paragraph{Societal Impact.} Our work focuses on improving the
video generation in technical aspects and
is not specifically designed for any malicious uses. This
being said, we do see that the method could be potentially
extended into controversial applications such as generating
fake videos. Therefore, we believe that the synthesized videos should present themselves as synthetic.
{
    \small
    \bibliographystyle{ieeenat_fullname}
    \bibliography{main}
}

\maketitlesupplementary
\setcounter{page}{1}

\begin{figure*}[h]
	\centering
	\includegraphics[width=0.99\textwidth]{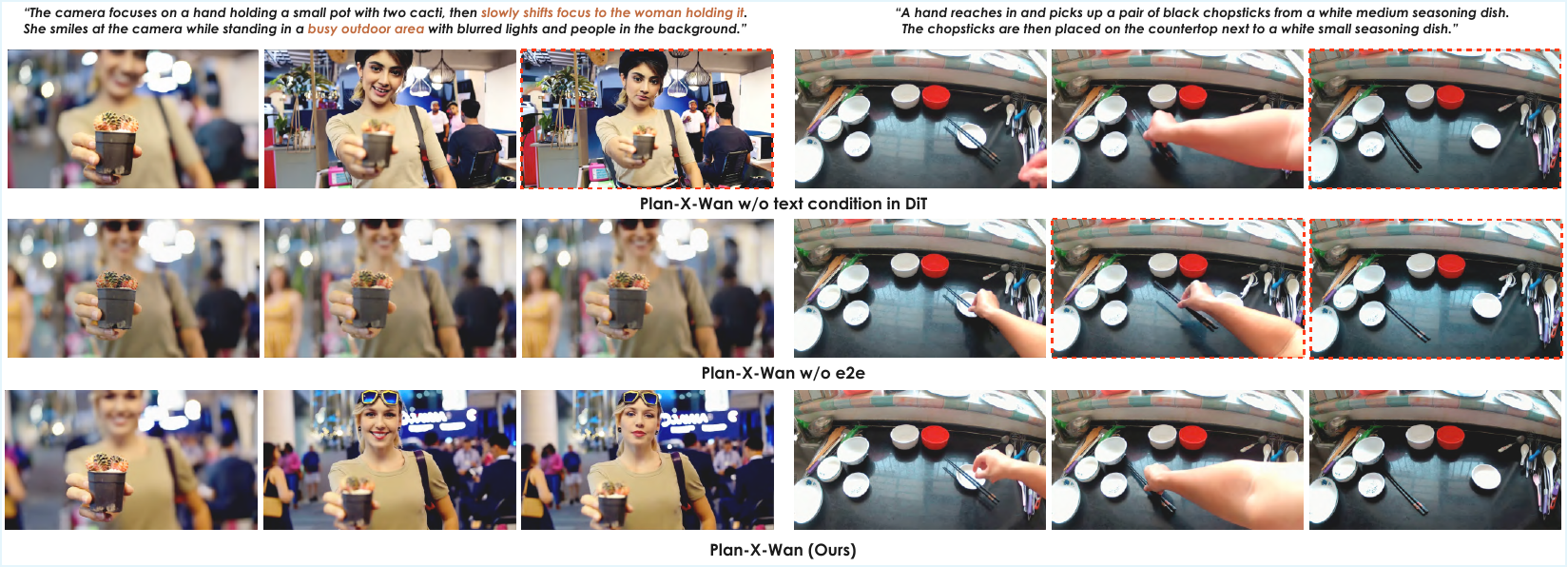}
	\caption{\textbf{Additional qualitative ablation.} Removing text conditioning in the DiT impairs visual quality (e.g., introducing oil-painting–like textures) and temporal consistency, whereas training \papername{} without joint end-to-end fine-tuning reduces robustness to semantic planning noise (for example, a white peeler incorrectly appears next to the white bowl holding the chopsticks in the second example).}
	\label{fig:more_aba}
\end{figure*}

\begin{figure*}[t!]
    \centering
    \begin{tcolorbox}[colback=gray!5,colframe=gray!50,title=\textbf{Semantic Planner Training Data Format}, width=\linewidth]
    \begin{lstlisting}[basicstyle=\small\ttfamily]
conversation = [
    {
    "role": "system",
    "content": "You are a helpful assistant."
    },
    {
    "role": "user",
    "content": f"Given the initial frame description: {image_description}, " +
             f"generate the subsequent {F-1} frames of the initial frame " +
             f"based on the instruction: {video_prompt}.\n<image>"
    },
    {
    "role": "assistant",
    "content": "<im_start><S2><image><im_end>"*(F-1)
    }
]
    \end{lstlisting}
    \end{tcolorbox}
    \caption{Semantic Planner Training Data Format.}
    \label{fig:planner_data_format}
\end{figure*}

In the supplementary material, we provide additional implementation training details (Section~\ref{sec:sup_imp_details}). In Section~\ref{sec:eval}, we further elaborate on our MLLM-based evaluation and user study setup.
We also ablate the impact of end-to-end training in Section~\ref{sec:more_aba}, and present representative failure cases to illustrate the current limitations of our framework (Section~\ref{sec:failure}).







\section{More Implementation Details}
\label{sec:sup_imp_details}
\subsection{Semantic Planner Training}
We format each training clip as a conversation, as illustrated in Figure~\ref{fig:planner_data_format}.

For frame representation, we use TA-Tok~\cite{han2025tar} tokens, which are trained at three adaptive spatial scales with token counts $\{729, 169, 81\}$ corresponding to scales $\{0, 1, 2\}$ from fine to coarse.
After applying the chat template, we replace the placeholder \texttt{<image>} token in the user prompt with 729 TA-Tok tokens representing the initial frame, preserving fine-grained perceptual details.
In the assistant response, each \texttt{<image>} placeholder token is replaced by 81 TA-Tok tokens at the coarsest pooling scale.
The special tokens \texttt{<im\_start>} and \texttt{<im\_end>} mark the boundaries of each image token sequence, while \texttt{<S2>} indicates the pooling scale and $F$ denotes the number of target video frames.
During training, we compute the cross-entropy loss over the TA-Tok tokens in the assistant’s response.

To enable hybrid text-to-video (T2V) and image-to-video (I2V) capabilities, we randomly zero out the TA-Tok embeddings in the user prompt with probability $50\%$.

\begin{figure*}[t]
	\centering
	\includegraphics[width=0.99\textwidth]{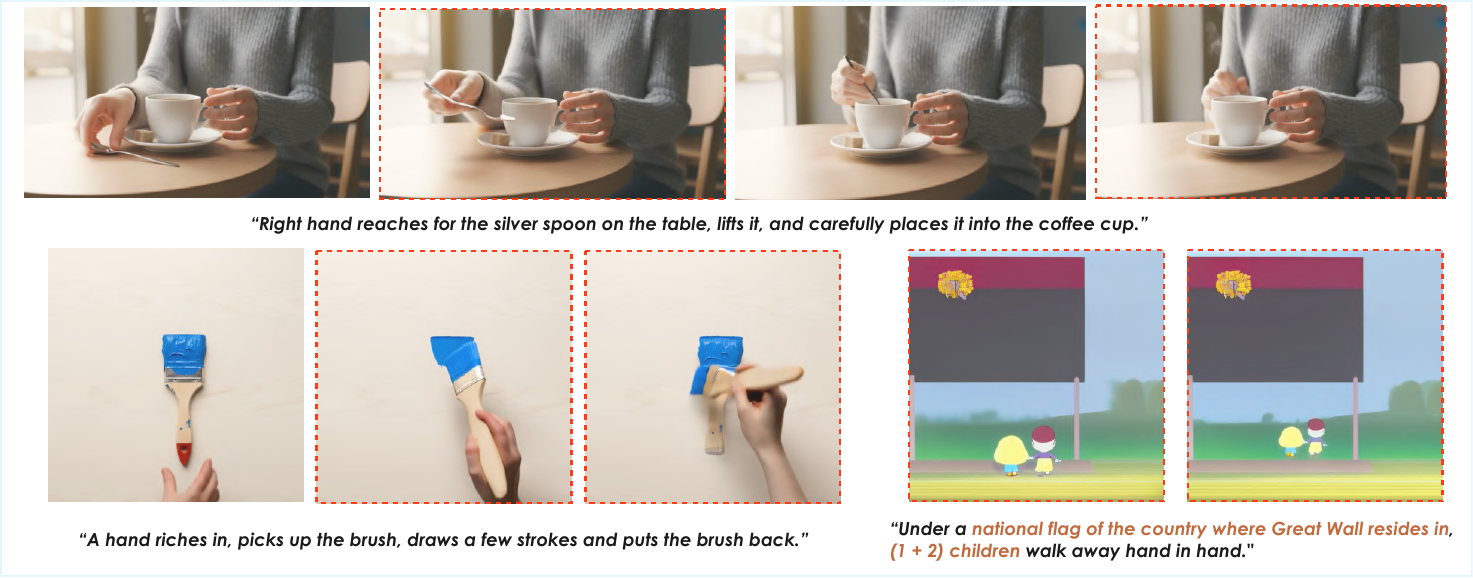}
	\caption{\textbf{Failure cases.} Top row: the spoon undergoes non-rigid morphing and eventually pops out of the coffee cup.
Bottom left: the red paint on the brush disappears.
Bottom right: the flag and the number of children do not match the prompt description. }
	\label{fig:failure}
\end{figure*}

\subsection{Semantics Guidance in DiT}
To inject semantic guidance into the DiT model, we replicate the original text branch into a semantic branch.
Concretely, we introduce new key and value projection layers for the semantic tokens, initialized from the corresponding text projection weights.

For Seedance~\cite{gao2025seedance} using MMDiT architectures, we directly append the keys $K_{\text{sem}}$ and values $V_{\text{sem}}$ derived from the semantic tokens to the existing text keys $K_{\text{text}}$ and values $V_{\text{text}}$:
\begin{equation}
    K = [K_{\text{text}} \parallel K_{\text{sem}}], \quad
    V = [V_{\text{text}} \parallel V_{\text{sem}}],
\end{equation}
where $\parallel$ denotes concatenation along the sequence dimension.

For Wan2.2~\cite{wan2025wan}, which uses cross-attention for text conditioning and does not apply RoPE in the original text cross-attention, we compute the attention logits for text and semantic tokens separately.
Let $Q$ denote the video query features.
We first compute
\begin{equation}
    L_{\text{text}} = \frac{Q K_{\text{text}}^T}{\sqrt{d}}, \quad L_{\text{sem}} = \frac{\tilde{Q} \tilde{K}_{\text{sem}}^T}{\sqrt{d}}
\end{equation}
We then concatenate these logits and apply the softmax function to compute the final attention output:
\begin{equation}
    \text{Attention} = \text{softmax}([L_{\text{text}} \parallel L_{\text{sem}}]) [V_{\text{text}} \parallel V_{\text{sem}}].
\end{equation}



\subsection{End-to-End Training.}
In the final stage, we jointly fine-tune both the semantic planner and the DiT model in an end-to-end manner.
Similar to BLIP3o-NEXT~\cite{chen2025blip3onextfrontiernativeimage}, instead of generating discrete tokens and querying the codebook embeddings for DiT conditioning, we directly project and inject the last hidden states (before the semantic planner’s linear prediction head) as the DiT semantic conditioning.
This design enables stable training with differentiable gradients and mitigates representation collapse caused by quantization.
For optimization, we use a weighted combination of the diffusion loss and the TA-Tok prediction loss, with weights of $1.0$ and $0.1$, respectively.

\section{Evaluation}
\label{sec:eval}

\paragraph{MLLM Evaluation Prompts.}
We utilize the SOTA multimodal LLM, Gemini-2.5~\cite{comanici2025gemini}, to evaluate the generated videos in multiple dimensions. The detailed system prompt is provided in Figure~\ref{fig:gemini_prompt}.

\begin{figure*}[h!]
    \centering
    \begin{tcolorbox}[colback=gray!5,colframe=gray!50,title=\textbf{Gemini Evaluation System Prompt}, width=\linewidth]
    \begin{lstlisting}[basicstyle=\fontsize{6pt}{7pt}\selectfont\ttfamily]
You are an expert video evaluator. Your task is to analyze a generated video based on a text prompt and an initial reference image.
Please evaluate the video on the following aspects. For each metric, provide a score between 0.0 and 1.0 and a brief "analysis"
 (a one-sentence justification for your score).
**Prompt:** "{prompt}"
**Initial Image:** [The image provided as the first frame]
---
## A. Prompt & Context Alignment
(How well the video matches the prompt and the initial image)
1. **Interaction Fidelity** (0.0, 0.1,...,1.0):
  * If the prompt describes an interaction, how accurately is it depicted?
  * **Analyze 3 components:**
    1.  **Interacting Objects:** Does the video show the *correct objects* (from the image) performing the interaction?
    2.  **Interaction Movement:** Is the *specific action* or movement itself correct (e.g., "pushing" vs. "lifting")?
    3.  **Interaction Outcome:** Is the *final state* or result of the interaction correct (e.g., "the vase falls off")?
* **Rating:**
    * 1.0 = Perfectly accurate. All 3 components (objects, movement, outcome) are correct. **(Or no interaction was described in the prompt).**
    * ...
    * 0.5 = Partially correct (e.g., correct objects but wrong movement/outcome).
    * ...
    * 0.0 = Fundamentally incorrect (e.g., wrong objects interacting or a completely different action).
2. **Prompt Completeness (Recall)** (0.0, 0.1,...,1.0):
    * Of all motions described *in the prompt*, what percentage are successfully completed *in the video*?
    * 1.0 = All prompted motions are fully completed.
    * ...
    * 0.0 = No prompted motions are completed.
3. **Motion Fidelity (Precision)** (0.0, 0.1,...,1.0):
    * Of all motions observed *in the video*, what percentage were *also described* in the prompt? (This measures motion hallucination).
    * 1.0 = No unprompted/hallucinated motion.
    * ...
    * 0.0 = All motion was unprompted/hallucinated.
4. **Object Consistency** (0.0, 0.1,...,1.0):
    * **Appearance:** Do objects from the initial image maintain a consistent appearance (shape, color, texture), or do they get corrupted/changed?
    * **Persistence:** Do all significant objects from the initial image remain present, and does the video avoid introducing new, duplicated objects excepts for hands performing the prompted actions?
    * 1.0 = Perfectly consistent: objects maintain their appearance, no objects disappear, and no new objects are introduced.
    * ...
    * 0.0 = Severely inconsistent: objects are corrupted, disappear, or new ones appear.
## B. Technical Quality
(The overall quality of the video file and its motion)
1. **Motion Naturalness** (0.0, 0.1,...,1.0):
    * How natural and realistic are the movements in the video?
    * Considers if motions are jittery, unnaturally fast/slow, or physically implausible.
    * 1.0 = Completely natural, realistic motion.
    * ...
    * 0.0 = Completely unnatural, broken, or incoherent motion.
2. **Visual Quality** (0.0, 0.1,...,1.0):
    * Visual quality including resolution, clarity, artifacts, stability.
    * Considers compression artifacts, blurriness, distortions, frame consistency.
    * 1.0 = High visual quality, clear and stable.
    * ...
    * 0.0 = Poor visual quality with major artifacts.
---
Please respond in this exact JSON format. Do not include any other text before or after the JSON block:
{{
    "prompt_context_alignment": {{
        "interaction_fidelity": {{
            "analysis": "Brief justification for interaction accuracy score.",
            "score": X.X
        }},
        "prompt_completeness_recall": {{
            "analysis": "Brief justification for prompt completeness score.",
            "score": X.X
        }},
        "motion_fidelity_precision": {{
            "analysis": "Brief justification for motion fidelity (hallucination) score.",
            "score": X.X
        }},
        "object_consistency": {{
            "analysis": "Brief justification for object consistency score.",
            "score": X.X
        }}
    }},
    "technical_quality": {{
        "motion_naturalness": {{
            "analysis": "Brief justification for motion naturalness score.",
            "score": X.X
        }},
        "visual_quality": {{
            "analysis": "Brief justification for visual quality score.",
            "score": X.X
        }}
    }}
}}
    \end{lstlisting}
    \end{tcolorbox}
    \caption{The system prompt used for Gemini-based evaluation.}
    \label{fig:gemini_prompt}
\end{figure*}

\paragraph{Human Evaluation Settings.}
We randomly select 20 samples from our test set for each model variant.
Each video is evaluated by 16 independent human raters.
We ask the raters to choose the top two videos based on overall quality and prompt alignment.
We then count how many times each result is selected as one of the top two and report these counts in Table~1 of the main paper.
Finally, we normalize the counts by the total number of selections to obtain the reported percentages.


\begin{table*}[ht]
	\centering
	\small
	\setlength{\tabcolsep}{7pt}
	\renewcommand{\arraystretch}{1.15}
    \vspace{-2mm}
	\caption{\textbf{Additional quantitative ablation.} We denote the best results in \textbf{bold}. }
	\label{tab:more_aba}
    \resizebox{0.9\textwidth}{!}{
	\begin{tabular}{lcccccc}
		\toprule
Method  & Accuracy $\uparrow$ & Completeness$\uparrow$ & Fidelity $\uparrow$ & Consistency $\uparrow$ & Naturalness $\uparrow$ & Visual $\uparrow$ \\
\midrule
\papername{}-Wan w/o text & 0.6533 & 0.7044 & 0.8133 & 0.8622 & 0.8444 & 0.9156 \\

\papername{}-Wan w/o e2e   & 0.7256 & 0.7535 & 0.8977 & 0.9767 & 0.9023 & 0.9682 \\

\midrule
\papername{}-Wan   & \textbf{0.7816} & \textbf{0.8263} & \textbf{0.9500} & \textbf{0.9816} & \textbf{0.9394} & \textbf{0.9737}  \\
\bottomrule
\end{tabular}
}
\vspace{-5mm}
\end{table*}

\section{Additional Ablations}
\label{sec:more_aba}
We conduct additional ablations to study the impact of both end-to-end joint training and text conditioning in the DiT.
As illustrated in Figure~\ref{fig:more_aba}, end-to-end training improves prompt-following behavior, while retaining the text-conditioning branch in the DiT helps preserve the pretrained visual quality. These trends are further confirmed quantitatively in Table~\ref{tab:more_aba}.

\section{Limitations}
\label{sec:failure}
In Figure~\ref{fig:failure}, we present several representative failure cases.
Since our framework relies on a pretrained DiT model for visual realization, it can still exhibit artifacts in physical fidelity and visual consistency.
Additionally, our semantic planner is trained on approximately 4.5M text–video pairs only, and thus lacks strong abstract reasoning and common-sense intelligence, as illustrated in the last row.

\end{document}